\documentclass[runningheads]{llncs}
\usepackage[T1]{fontenc}
\usepackage{graphicx} % Required for inserting images
\usepackage{hyperref}
\usepackage{float}
\usepackage{url}
\usepackage{array} 
\usepackage{amsmath}
\usepackage[misc]{ifsym}
\begin{document}
\title{ShizishanGPT: An Agricultural Large Language Model Integrating Tools and Resources}
\titlerunning{ShizishanGPT}
\author{Shuting Yang\inst{1}, Zehui Liu\inst{1},Wolfgang Mayer , Ningpei Ding,}
\author{Shuting Yang\textsuperscript{1\#}
\and Zehui Liu\textsuperscript{1\#} 
\and Wolfgang Mayer\textsuperscript{2}
\and Ningpei Ding\textsuperscript{1}
\and Ying Wang\textsuperscript{1}
\and Yu Huang\textsuperscript{1}\textsuperscript{\Letter}
\and Pengfei Wu\textsuperscript{1}\textsuperscript{\Letter} 
\and Wanli Li\textsuperscript{1}\textsuperscript{\Letter} 
\and Lin Li\textsuperscript{3}\textsuperscript{\Letter} 
\and Hong-Yu Zhang\textsuperscript{1}\textsuperscript{\Letter}
\and Zaiwen Feng\textsuperscript{1}\textsuperscript{\Letter}
\thanks{Correspondences:yhuang,chriswpf,liwanli,hzaulilin,zhy630,Zaiwen.Feng@mail.hzau.edu.cn.
\textsuperscript{\#}These authors contribute equally to this work.}}
\authorrunning{S. Yang et al.}
\institute{College of Informatics, Huazhong Agricultural University, Wuhan, 430070, China\and
Industrial AI Research Centre, University of South Australia, Mawson Lakes, SA, 5095, Australia\and
College of Plant Science and Technology, Huazhong Agricultural University, Wuhan, 430070, China}
\maketitle              % typeset the header of the contribution
\begin{abstract}
Recent developments in large language models (LLMs) have led to significant improvements in intelligent dialogue systems' ability to handle complex inquiries. However, current LLMs still exhibit limitations in specialized domain knowledge, particularly in technical fields such as agriculture. To address this problem, we propose ShizishanGPT, an intelligent question answering system for agriculture based on the Retrieval Augmented Generation (RAG) framework and agent architecture. ShizishanGPT consists of five key modules: including a generic GPT-4 based module for answering general questions; a search engine module that compensates for the problem that the large language model's own knowledge cannot be updated in a timely manner; an agricultural knowledge graph module for providing domain facts; a retrieval module which uses RAG to supplement domain knowledge; and an agricultural agent module, which invokes specialized models for crop phenotype prediction, gene expression analysis, and so on. We evaluated the ShizishanGPT using a dataset containing 100 agricultural questions specially designed for this study. The experimental results show that the tool significantly outperforms general LLMs as it provides more accurate and detailed answers due to its modular design and integration of different domain knowledge sources. Our source code, dataset, and model weights are publicly available at \url{https://github.com/Zaiwen/CropGPT}.


\end{abstract}

\section{Introduction}
\begin{figure}[H]
    \centering
    \includegraphics[width=1.0\textwidth]{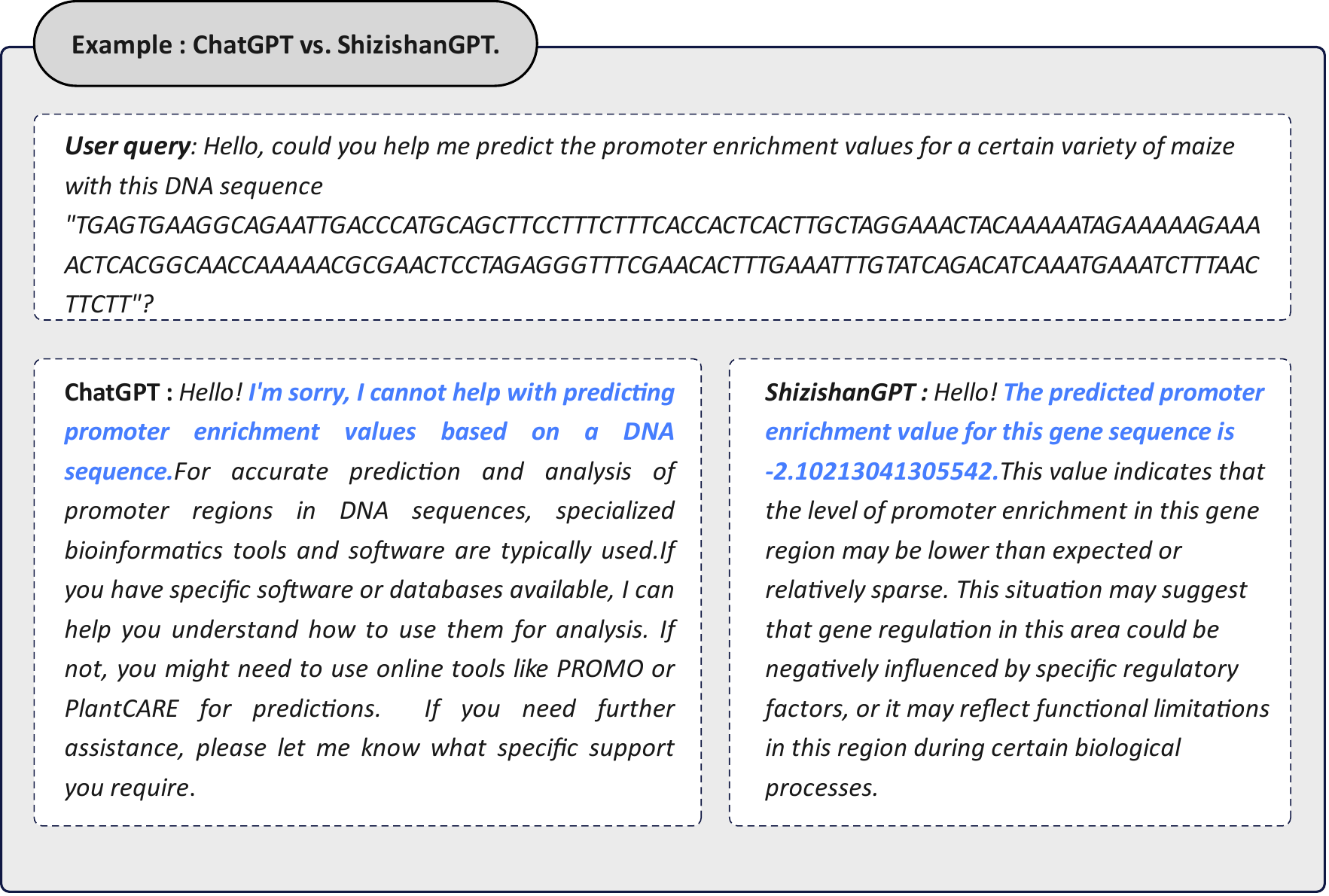}
    \caption{Example of ChatGPT versus ShizishanGPT in predicting maize gene promoter enrichment values }
    \label{fig:example}
\end{figure}

Recently, the advancement of large model technology has marked a significant milestone in the field of artificial intelligence, showcasing impressive performance across various industry sectorss~\cite{ref_proc1}. These models, trained on vast datasets, have developed a remarkable ability to understand and generate human-like language, which has led to their rapid adoption in sectors such as healthcare and education.
Despite their widespread utility, the adaptation of language models for specialized domains, particularly in agriculture and more specifically in the crop domain, remains insufficiently developed~\cite{ref_proc2}. This limitation significantly impacts the application of such models in agricultural practices, including the utilization of advanced genetic tools and the retrieval of specialized crop-related knowledge, which are critical for precision farming and enhanced crop management. To address this problem, this paper proposes an intelligent question answering system for agriculture based on the Retrieval Augmented Generation (RAG) framework and agent architecture (ShizishanGPT). 
The system uses a large language model to answer questions related to the cultivation of crops (especially maize) and helps users to solve problems by providing accurate answers through access to relevant crop tools. For individuals who lack expertise in crops, the system empowers smart agriculture by integrating specific functional requirements such as crop cultivation, enabling them to excel in knowledge-based tasks within their area of expertise~\cite{ref_proc3}. 
 For instance, in scenarios where a user needs to predict promoter enrichment 
 values (a key indicator of gene expression potential in specific DNA segments of the promoter region) ShizishanGPT leverages its 
 specialized modules to invoke external models that perform detailed calculations and return precise predictions. This capability contrasts sharply with general-purpose models like ChatGPT-4~\cite{ref_proc4}, which, lacking domain-specific training and integration capabilities, often fail to provide accurate or relevant responses in the agricultural context. 
 Figure~\ref{fig:example} shows the contrasting responses from "ChatGPT" and "ShizishanGPT" to a question about predicting promoter enrichment values for a maize DNA sequence. highlighting ShizishanGPT's strength in solving specific problems in the agricultural domain and providing accurate and credible agricultural knowledge.

 In this paper, the architectural underpinnings of ShizishanGPT include several key components, which include an Agricultural Knowledge Graph that maps out critical data and relationships within the agricultural domain; a Vector Knowledge Base  that contains extensive papers in the domain related to Crop; External tools which contain Gene Sequence Prediction tool that aid in the analysis of genetic data and Maize Phenotype Prediction tools~\cite{ref_proc5}that help anticipate crop characteristics based on genetic and environmental factors. We evaluate our system using a dataset containing 100 question-answer pairs in the field of agriculture, specially designed for this study, and demonstrate the effectiveness of the proposed system. The paper is structured as follows: the introduction provides background information on the research objectives; the second section reviews related research and technologies; the third section details the design and implementation of the system; the experimental section discusses the experimental validation and investigates the contributions of each element in an ablation study; potential limitations of the study are discussed in the subsequent section; the conclusion summarizes the contributions.
 
\section{Related Works}
This section reviews related works on enhancing large language models (LLMs), focusing on the integration of Retrieval Augmented Generation (RAG) and agent-based frameworks, the use of knowledge graph-enhanced models, and the application of external tools. It also discusses the implementation of knowledge graphs in agriculture, showcasing how these technologies improve the accuracy and utility of large models in complex tasks.

\subsection{RAG and Agent-Based Frameworks for LLM Enhancement}

Since the advent of LLMs like ChatGPT, these models have excelled in a variety of language tasks~\cite{ref_proc6} but still face challenges such as hallucinations~\cite{ref_proc7}, outdated knowledge, and data constraints that affect their reliability, especially in tasks requiring extensive and precise knowledge like open-domain question answering and commonsense reasoning.
The implementation of RAG has provided a pathway to address some of these limitations by enhancing the response accuracy of LLMs in knowledge-intensive scenarios. The concept of Retrieval-Augmented Generation, initially
proposed by Lewis et al~\cite{ref_proc8}. RAG has significantly enhanced the effectiveness of LLM responses by providing contextual grounding from existing documents during the generation process~\cite{ref_proc9}. RAG merges
the retrieval and answering processes, which improves the ability to effectively collect knowledge,
extract useful information, and generate answers. Additionally, the Reasoning and Action (ReAct)~\cite{ref_proc10} framework aids LLMs in managing complex tasks by integrating multiple reasoning and action steps into the problem-solving process, thus leading to more precise outputs~\cite{ref_proc11}.

ShizishanGPT leverages both Retrieval RAG and ReAct to boost its performance in complex agricultural tasks. Specifically enhanced by training on a corpus of over a thousand scholarly articles, RAG improves the factual accuracy and relevance of responses. Meanwhile, ReAct facilitates the handling of intricate problem chains through structured reasoning and action sequences. This combination enables ShizishanGPT not only to provide data-driven, precise answers but also to continually adapt to the evolving needs of the agricultural sector, significantly enhancing the system’s utility and reliability in agricultural decision support.

\subsection{Knowledge Graph Enhanced Large Models}

 LLMs have become an important technology in modern AI, especially in the field of Natural Language Processing (NLP), and are widely used in a variety of language understanding and generation tasks. These models, such as OpenAI's GPT family, Google's BERT~\cite{ref_proc12} and T5~\cite{ref_proc13}, Facebook's BART~\cite{ref_proc14}, LLama~\cite{ref_proc15}, learn rich linguistic, contextual, and knowledge information by pre-training on large-scale datasets, enabling them to perform various linguistic tasks without explicit task instructions. In particular, LLMs have demonstrated exceptional capabilities in question-answering systems. These models can provide exhaustive answers and generate innovative answers in specific situations by learning from data, facts and world knowledge. For example, GPT-3~\cite{ref_proc16}, with its large model structure and training data, can answer complex questions without external database support. 

To further improve answer quality and accuracy, knowledge graphs and external tools are becoming increasingly important in modern LLM-based QA systems. For example, in December 2023, a research team from China Agricultural University (CAU) released ‘Shennong Big Model 1.0’~\cite{ref_url1}, an industry big model applied to the agricultural field. The model is trained from a large amount of high-quality agricultural knowledge data, including more than 10 million agricultural knowledge mapping data, more than 50 million modern agricultural production data and more than 20,000 agricultural books. Based on the traditional big model technology architecture, it integrates knowledge graph, vector database and other technologies to improve the accuracy of the answers.

\subsection{External Tools Enhanced Large Models}

Tool-enhanced Large Language Models (LLM) and agent-based large language models frameworks have received much attention for their application in agriculture. The Tool Augmented Large Language Model framework utilises information from external tools to enhance the capabilities of the language model.~\cite{ref_proc18} In the agricultural domain, these external tools can include maize phenotype prediction models, knowledge graphs in the agricultural domain as well as related literature, Google search engine, etc. By integrating with these tools, the language model can understand and solve agriculture-related problems more accurately. For example, Qing et al~\cite{ref_proc19} designed a pest identification system that integrates a large language model with an external computer vision model, effectively combining it with other small external models to enhance the text comprehension and generation capabilities of the large model. Balaguer et al~\cite{ref_proc20} used retrieval-enhanced generation~\cite{ref_proc21} and fine-tuning techniques to improve the model's responsiveness to region-specific problems. HuggingGPT~ \cite{ref_proc22} combines OpenAI's GPT-4~\cite{ref_proc4} with the Hugging Face model library. It allows GPT-4 to act as a central manager, coordinating and invoking hundreds of small models as external tools to solve specific tasks. This multi-model collaboration mechanism improves the adaptability of the system, allowing it to take advantage of the strengths of different models to solve a variety of complex problems.
The agent-based language modelling framework handles requests to language models through agents to improve the efficiency and performance of the system.
\subsection{Knowledge Graphs in Agriculture}
In agriculture, researchers are harnessing knowledge graphs to boost production efficiency and sustainability. These graphs amalgamate diverse agricultural data, aiding in decision-making and providing precise management solutions. Knowledge graphs, with their structured representation of knowledge, enhance the interpretability of large language models used in their training~\cite{ref_proc23}. Therefore, introducing knowledge graphs in the training process of large language models helps to improve the interpretability of large language models. For example, Rezayi et al~\cite{ref_proc24} proposed the AgriBERT model for matching food and nutrients. The model, based on the structure of the BERT language model, was pre-trained on a corpus dataset of academic journals and fine-tuned by augmenting the answers with a map of agricultural expertise, and the results showed a significant improvement in the matching ability of the model. AgriKG~\cite{ref_proc25}, a comprehensive agricultural knowledge graph utilized in this study, employs natural language processing to extract and connect entities and their relationships from extensive agricultural data. This graph supports applications such as entity retrieval and intelligent QA. Leveraging AgriKG, this study develops an intelligent QA system using large-scale language models. This system integrates AgriKG's data with user queries to deliver effective QA services.

\section{System Design}
\subsection{Design of Prompt}
In order to ensure that the model is able to generate relevant and accurate responses to the complex needs of the agricultural domain, special attention has been invested in the design of the system's prompts. When processing and analyzing agricultural data, well-designed prompts are essential to guide the model in extracting and utilizing key information. Therefore, we designed the hint generation mechanism in our system to dynamically adjust to the specific content of the query, thus improving the accuracy and efficiency of information retrieval.
\subsubsection{Prompts Mechanisms:}
In the system, the design of prompts is based on the automatic analysis of user queries, from which key information words and action words are extracted, such as "corn," "yield," or "predict." These terms are used to construct structured prompts tailored to specific queries, guiding data retrieval and analysis. For example, for a query about the predicted phenotype value of a specific DNA sequence of maize in the Huazhong region, the prompt might be "Analyze the impact of a certain DNA sequence on maize growth rate under the climatic conditions of the Huazhong region. " Based on these prompts, the system invokes relevant knowledge graphs, databases, and prediction models, and selects appropriate analytical tools to optimize query processing, ensuring accurate and relevant results.

\subsubsection{The Role of Prompts in System Architecture:}
Prompts act as a bridge between the user query and the system's comprehensive database, ensuring that the model's answers are not only relevant but also rich in domain-specific insights. By structuring key descriptors and verbs in the prompts, the system is able to utilise its resources more efficiently, therefore, the design of the prompts has a direct impact on the efficiency with which the system can process a particular query, speeding up the response and improving accuracy by reducing the amount of data to relevant information, this approach enhances the system's ability to provide the user with solutions to a particular problem.
\subsection{Workflow of System}
The general workflow of the system is as follows: When a user inputs a question, the system analyzes the question and selects the most relevant type of problem description from its various functional modules. It then calls the appropriate tools to complete the task. After receiving feedback, the large language model dynamically adjusts the next step of the plan based on the feedback. Finally, it summarizes based on the results and its own knowledge, ultimately providing the user with a clear and useful answer. Our system employs the ReAct architecture; if the agent's first tool selection fails, it reanalyzes the problem based on the returned results, chooses other appropriate tools for iterative execution, can also decompose the problem into sub-problems, calls tools to obtain feedback, and then executes the next step based on the feedback.

\subsection{System Architecture}
\begin{figure}[htp]
    \centering
   
    \includegraphics[width=1.0\textwidth]{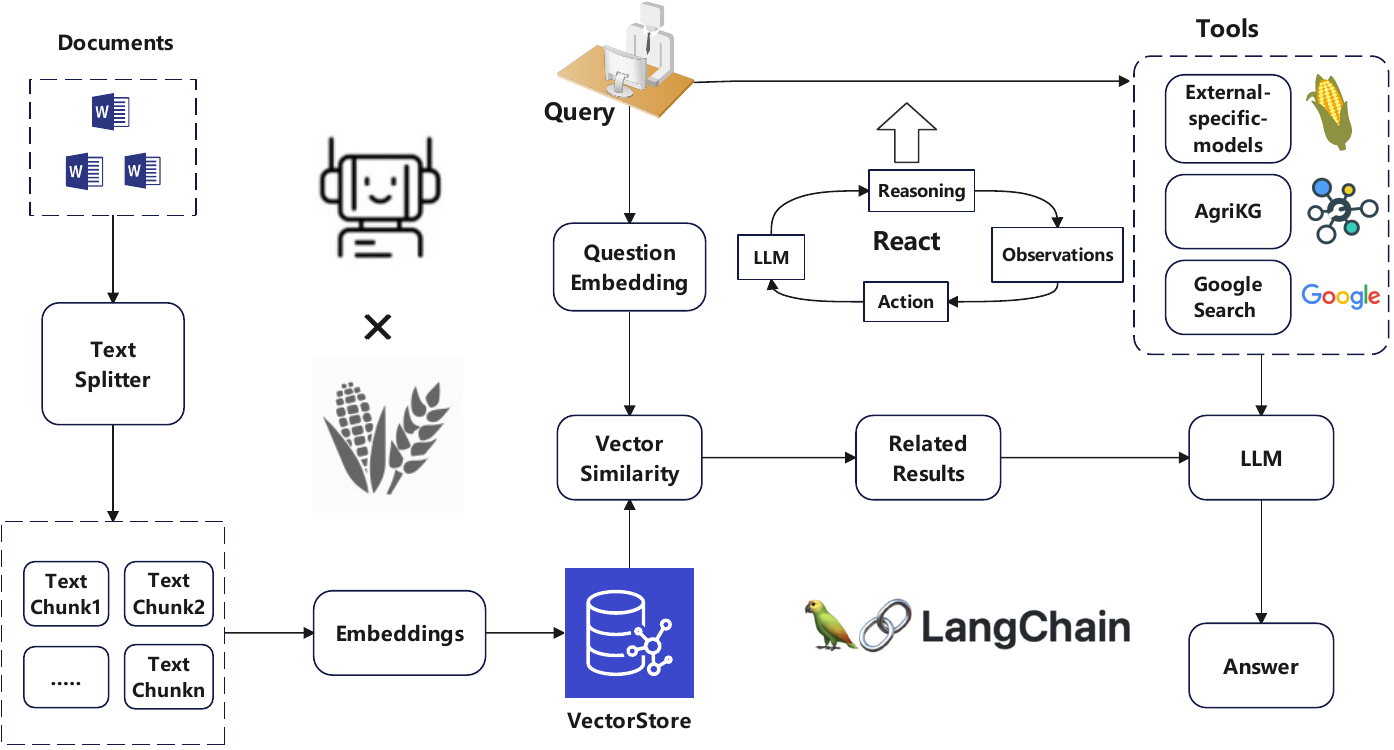}
    \caption{Detailed Architecture Diagram of the Question Answering Pipeline}
    \label{fig:detailed pipeline}
\end{figure}

We propose an intelligent QA system for agricultural knowledge based on the RAG framework and agent architecture, which adopts an integrated architecture that combines advanced artificial intelligence techniques and data management technologies, aiming to provide efficient and accurate QA services. The system comprises multiple modules optimized for a specific function to ensure comprehensive and accurate question answering, shown in Figure~\ref{fig:detailed pipeline}. Specifically, the system utilizes a large pre-trained language model to handle generic queries and a search engine module to access the latest agricultural data and research in real time. In addition, the system is embedded with a search vector database of specialized literature to support accurate answers to in-depth domain questions. Using the AgriKG knowledge graph, the system can provide detailed answers to complex questions about crop types, suitable growing conditions, etc. The system also integrates several small prediction models, such as maize phenotype prediction (Resnet model architecture) and rice gene promoter enrichment value prediction, to provide accurate agricultural decision support. The synergistic work of these modules, managed through a unified user interface, ensures the simplicity and efficiency of the system's operation, enabling users to get instant and accurate answers to their natural language queries, thus greatly improving the intelligence of agricultural production and management.

\subsection{Modules in the Architecture}
The modules are as follows: 

\subsubsection{Problem analysis module: }The Generic Module is based on large pre-trained language models and focuses on generic questions, such as queries about the basics of agriculture. This module parses and answers questions using trained NLP models chosen for their broad knowledge base and strong language understanding. The generic module first analyzes the user query to determine its nature and whether additional knowledge sources are required.

\subsubsection{Search module:}The module can retrieve the latest agriculture-related information from external search engines, such as Google, and inject the resulting information into the LLM prompt for answer generation. It aggregates and filters search results from multiple sources, extracts key content from web pages, articles, and reports, and presents answers in summary form. By combining the information obtained from the search engine with the knowledge from other modules within the system, the search engine module provides users with rich and cutting-edge answers to agricultural knowledge, especially for complex questions.

\subsubsection{Retrieval module:}
This module leverages advanced natural language processing techniques by utilizing RAG~\cite{ref_proc21} technology to bolster our QA system's capabilities in a two-step process: retrieval and generation. Initially, a robust vector database searches through over a thousand specialized agricultural documents, including texts on crop breeding, disease control, and recent agricultural innovations, to find those most pertinent to the user's query. Once relevant literature is retrieved, a generative model uses this information to craft precise responses. This extensive repository not only ensures rapid retrieval of relevant documents but also provides essential context, greatly enhancing the QA system's ability to handle specialized queries in agriculture. The system quickly retrieves the document(s) closest to the user query. These retrieved documents provide critical contextual information that lays the foundation for subsequent answer generation as they significantly improve the performance of the QA system on specialized agricultural questions.
    
\subsubsection{Knowledge Graph module:}
The Knowledge Graph module utilizes the AgriKG Knowledge Graph~\cite{ref_proc4} to inject structured agricultural knowledge into the system's answering process, enhancing the expertise and accuracy of the questions and answers. By pre-indexing information such as crop classification, climate suitability, and pest control in AgriKG, the module can quickly retrieve the relevant knowledge required for a question. It maps the crop names and species entered by the user to entities and relationships in the knowledge graph through an API interface, extracting information such as crop classification and suitable growing environments. The module can also recommend the best planting strategies and optimization suggestions based on the structured expert knowledge in the knowledge graph.

\subsubsection{External tool invocation module:}
This module integrates a variety of specialized predictive models about specific crops, including a maize phenotype prediction model, maize gene promoter enrichment value prediction model, crop growth prediction model, crop variety selection model, pest and disease monitoring and control model, agricultural weather warning model, agricultural technology consulting model, and crop market analysis model to answer crop-related questions. The module automatically selects the appropriate model based on the user's question and invokes the model to obtain prediction and analysis results. The module then combines these results with knowledge from other modules within the system to synthesize multi-faceted answers. Using this module, the system can accurately solve complex crop prediction and analysis problems and provide detailed agricultural recommendations to the user.

\section{Experiments}

\textbf{Dataset:}
Since there was no suitable QA dataset to fully test system performance, we did the work of building an agricultural QA dataset. Firstly, some representative problems are found from existing agricultural databases, academic papers, agricultural knowledge websites, etc. We consider crop cultivation, agricultural technology, pest control, agricultural processing and other aspects to ensure a comprehensive and diverse set of issues. After getting the seed questions for each aspect, we used ChatGPT's help to generate some related questions to expand our problem set. The quality of the dataset is then improved by manual screening to remove duplicate questions and irrelevant content. In order to ensure the accuracy and reliability of the answers, we get the correct answers to these questions from professional agricultural websites. Finally, a dataset containing 100 high-quality question answer pairs was constructed as the experimental benchmark to evaluate and test the performance of the system in the field of agricultural QA.

\subsection{Machine Scoring}
\textbf{Method}: We use BLEU, ROUGE, and GLEU to score standard and system-generated answers. For each question, a reference answer on a professional agricultural website is used as the benchmark for the answer generated by the evaluation system. The answers generated by each system were scored using the following evaluation indicators.

    \textit{BLEU}: Calculates the n-gram overlap between the system's generated answers and all reference answers, and calculates BLEU scores based on accuracy, recall, and n-gram matches. Relying too heavily on phrase matching in BLEU scoring is not advisable, as it may fail to consider contextual and semantic nuances. \textit{ROUGE}: Calculates the degree of n-gram overlap between the system generated answer and the reference answer, and calculates the ROUGE score based on n-gram matches of different lengths. \textit{GLEU}: The system evaluates the syntactic and semantic consistency between the generated answer and the reference answer, and calculates the GLEU score based on n-gram matches of different lengths. GLEU provides a more fine-grained assessment than BLEU, but is more computationally expensive.

A higher score indicates that the answers generated by the system are largely similar to the standard answers, while a lower score indicates a gap.

\subsubsection{Scoring Standard:}
%\noindent
\begin{equation}
\text{Composite Score} = \alpha \times \text{BLEUScore} + \beta \times \text{ROUGEScore} + \gamma \times \text{GLEUScore}
\end{equation}
Among them, $\alpha$, $\beta$ ,$\gamma$ are the weights of each evaluation index, and  $\alpha + \beta + \gamma = 1$, here we set 
 $\alpha$, $\beta$, $\gamma$ to 0.4, 0.4, 0.2 respectively. The following is the specific formula of each evaluation index:

\begin{equation} %\label{eq:bleuscore}
\text{BLEUScore} = \text{BP} \times \left( \frac{\sum_{n=1}^4 \text{BLEU}_n}{4} \right) \tag{2}
\end{equation}
Where BP is the short penalty factor and $BLEU_n$ represents the $BLEU$ fraction of an n-gram.
\begin{equation}
\text{BLEU}_n = \frac{\text{Count}\_{\text{clip}}(n)}{\text{Count}\_{\text{gen}}(n)} \tag{3}
\end{equation}
$Count\_clip(n)$ is the minimum number of N-grams that the system generates to match the reference answer. $Count\_gen(n)$ is the number of N-grams in the answer generated by the system.
\begin{equation}
\text{ROUGEScore}(\text{RS}) = (1 - \lambda) \times \text{ROUGE}_L + \lambda \times (\text{ROUGE}_{\text{SU}} \times \omega_1 + \text{ROUGE}_{\text{LS}} \times \omega_2) \tag{4}
\end{equation}
Where $\lambda$ is the long sentence penalty factor, $ROUGE_{SU}$ is the unordered ROUGE fraction, $ROUGE_{LS}$ is the ordered ROUGE fraction, $\omega_1$ and $\omega_2$ are the weights, and$\omega_1$ + $\omega_2$ = 1, where we set $\lambda$ to 0.1 and both $\omega_1$ and $\omega_2$ to 0.5.

\begin{equation}
\text{GLUEScore} = \frac{\sum_{n=1}^4 \text{GLUE}_n}{4} \tag{5}
\end{equation}

\begin{equation}
\text{GLEU}_n = \frac{2 \times \text{Count\_match}(n)}{\text{Count\_gen}(n) + \text{Count\_ref}(n)} \tag{6}
\end{equation}
$Count\_match(n)$ is the number of N-grams matched between the system generated answer and the reference answer.
Through the above formula, we can get a comprehensive score considering multiple evaluation indicators. Adjust the weights of \(\alpha\), \(\beta\), \(\gamma\), \(\delta\), \(\lambda\), \(\omega_1\), and \(\omega_2\), and weigh the importance of different evaluation indicators according to specific needs to get the final result.

\subsubsection{Result:}
\begin{figure}[htp]
    \centering
    \includegraphics[width=1.0\textwidth]{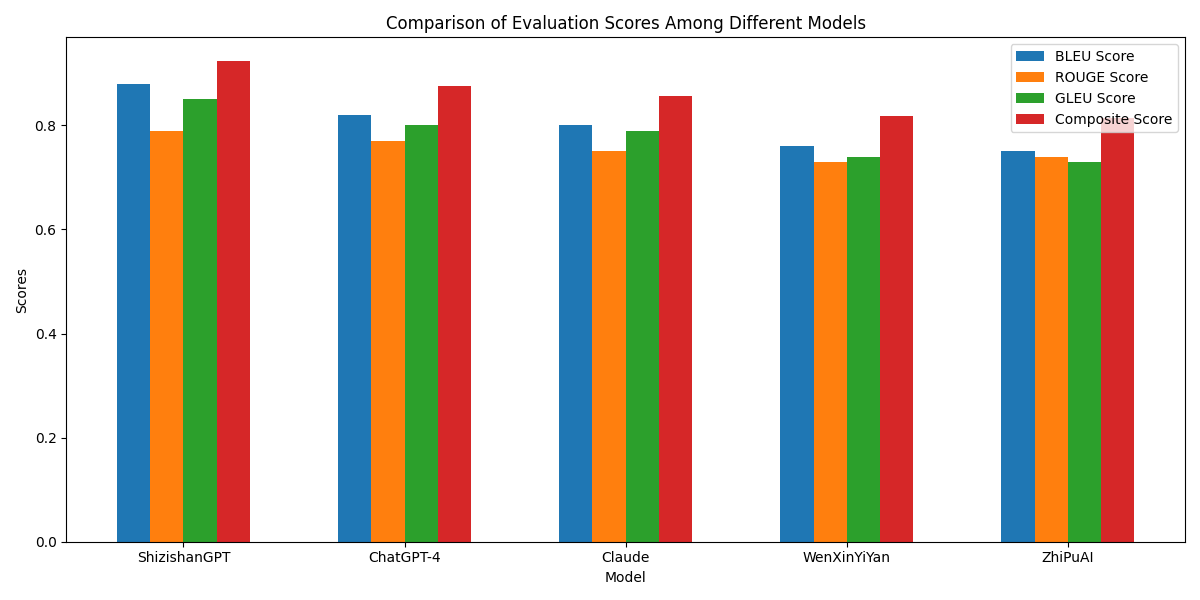}
    \caption{Comparative Analysis of Language Model Scores on BLEU, ROUGE, GLEU, and Composite Metrics}
    \label{fig:result}
\end{figure}

We evaluated the performance of the ShizishanGPT in the agricultural question answering task and compared it with several other models. The results of the comparison are shown in Figure~\ref{fig:result}. We set the standard answer score in the data set to 1.0, and it can be seen that ShizishanGPT significantly outperforms the other models in the BLUE and GLUE metrics. The performance on ROUGE metrics is similar to other models. This indicates that it outperforms ChatGPT and other models in terms of similarity and semantic consistency.

Table~\ref{tab:model_scores} presents a comparison of model scoring across different metrics, highlighting ShizishanGPT's superior performance with a BLEU score of 0.88, a ROUGE score of 0.79, and a GLEU score of 0.85, culminating in a composite score of 0.923. This outpaces other models such as ChatGPT-4, Claude, WenXinYiYan, and ZhiPuAI, which demonstrate varying degrees of effectiveness with ChatGPT-4 performing relatively well with a composite score of 0.876.ShizishanGPT can acquire knowledge that the model itself does not have from the literature vector library and knowledge graph, and deal with skills that other models do not have from external tools. This result shows that the ShizishanGPT has higher accuracy and reliability in intelligent question-answering in the agricultural field.

\begin{table}[H]
\centering
\caption{Comparison of model scoring}\label{tab:model_scores}
\begin{tabular}{|l|l|l|l|l|}
%\begin{tabular}{@{}lcccc@{}}  % l for left alignment, c for center
\hline
Model      & BLEU Score & ROUGE Score & GLEU Score & Composite Score \\ 
\hline
ShizishanGPT & 0.88       & 0.79        & 0.85       & 0.923           \\
ChatGPT-4  & 0.82       & 0.77        & 0.80       & 0.876           \\
Claude     & 0.80       & 0.75        & 0.79       & 0.857           \\
WenXinYiYan     & 0.76       & 0.73        & 0.74       & 0.818           \\
ZhiPuAI    & 0.75       & 0.74        & 0.73       & 0.815           \\
\hline
\end{tabular}
\end{table}

\subsection{Manual Scoring}

\textbf{Scoring Guidelines:}
In this study, we conducted a manual evaluation through several key steps to ensure the fairness and accuracy of the assessment. First, we selected evaluators with a background in agriculture, who specialize in assessing question-and-answer systems. The evaluators scored the answers generated by the models without knowing their sources.
The scoring criteria were divided into three categories: Accuracy, Professionalism, and Language Fluency. Each category’s specific standards were thoroughly explained during the training of evaluators to ensure consistency and accuracy in scoring. Each category had a maximum score of 10, and the final total score was calculated using the following weights: Accuracy accounted for 50\%, Professionalism for 30\%, and Language Fluency for 20\%. This weighting ensured a comprehensive assessment of the models' overall performance. To enhance the reliability of the evaluation, we implemented a cross-checking mechanism, ensuring that each answer was scored by at least three independent evaluators. This helped eliminate potential biases from individual scorers and also allowed us to verify the stability and fairness of the scoring system by comparing the consistency of scores among different evaluators.

\textbf{Scoring Criteria:}
Accuracy: Assesses how well the model answer corresponds to the requirements of the question and how correct the information is.
Professionalism: To see if the terms and concepts used in the answer are in line with standards and common sense in the agricultural field.
Language Fluency: Evaluate whether the answer is natural and grammatically correct.

\begin{table}[H]
\centering
\caption{Results on Agricultural QA Tasks}\label{tab:Evaluation}
\begin{tabular}{|l|l|l|l|l|}
\hline
Model Name &Accuracy& Professionalism & Language Fluency & Total Score \\
\hline
GPT4 & 9.2&9.3 & 9.5& 9.29\\
Claude & 9.1&9.3& 9.6 &9.26 \\
WenXinYiYan &8.9 & 9.4& 9.2&9.11 \\
ZhiPuAI &8.7 &9.1 &9.3 & 8.94\\
ShizishanGPT &9.8 & 9.5& 9.5& 9.65\\
\hline
\end{tabular}
\end{table}  

Table~\ref{tab:Evaluation}  shows the results of the experiment, 
 Our results indicate that ShizishanGPT outperformed other models with a composite score of 9.65, excelling particularly in accuracy (9.8) and professionalism (9.5). On the other hand, ZhiPuAI scored the lowest overall, with a total of 8.94, suggesting areas for improvement, especially in accuracy (8.7). Models like GPT4 and Claude showed competitive results with overall scores of 9.29 and 9.26, respectively, demonstrating strong language fluency capabilities. These findings highlight the significant potential of employing specialized language models in enhancing the precision and effectiveness of responses in agricultural domains.

\subsection{Ablation Experiments}
We conducted an ablation experiment to demonstrate each component's impact on our system. We disabled individual aspects of the system, such as RAG and external predictive models, and evaluated the system's performance again. We used the same QA dataset, metrics, and analysis procedure as in the previous experiment. Table~\ref{tab:ablation} shows the results of our experiment for the different configurations.

\begin{table}[H]
\centering
\caption{Results of the ablation study}\label{tab:ablation}
\begin{tabular}{|l|l|l|l|l|}
\hline
Configuration & Accuracy & Professionalism  & Language Fluency & Total Score \\
\hline
Full ShizishanGPT &9.8 &9.5 &9.5 &9.65 \\
- RAG &9.1 &9.2 &9.5 &9.21 \\
- Tools &9.4 &9.3 &9.4 &9.37 \\
- AgriKG &9.6 &9.4 &9.4 &9.50 \\
- Search &9.5 &9.3 &9.5 &9.44 \\
\hline
\end{tabular}
\end{table}  

 As can be seen from the data, the full configuration of ShizishanGPT demonstrated the highest overall performance with scores of 9.8 in Accuracy, 9.5 in Professionalism, and 9.1 in Language Fluency, culminating in a total score of 9.65. When individual components were incrementally removed, the model showed varying degrees of performance impact. Notably, the removal of AgriKG resulted in a slight improvement in Accuracy and Professionalism, indicating a robustness in the remaining components, with a new total score of 9.50. Conversely, the removal of the Retrieval-Augmented Generation (RAG) component resulted in the most significant drop in total score to 9.21, underscoring its critical role in enhancing model performance. Each configuration, whether '- Tools', '- Search', or others, exhibited distinct impacts on specific metrics, but consistently maintained high performance across all metrics, demonstrating the architecture's resilience and the effectiveness of its components.

\section{Limitations}

Although the results of our experiment are encouraging, it is important to discuss potential limitations.

%\begin{description}
    \textbf{Dataset limitations:} The dataset used in this article was generated by GPT-4 and may be affected by the quality of the questions due to limitations and biases in GPT-4's training data and question generation capabilities. Since the seed problem was found manually, the representativeness of the dataset may also have certain limitations.

    \textbf{Scope limitations:} In terms of universality and generalization, the research in this paper may only apply to a specific environment or scope of questions. The research in this article may apply only to the field of agriculture and only to specific crops and regions.

\section{Conclusion}
This study successfully integrates a knowledge graph-enhanced large-scale language model with an external tool module and a retrieval vector database module to construct a comprehensive intelligent QA system for agriculture. Experimental results show that the system exhibits good performance and accuracy in answering agriculture-related questions, and the response quality of the system is further enhanced by calling the external tool module and retrieval vector database module.

We have successfully demonstrated intelligent querying, reasoning, and retrieval of agricultural knowledge, providing more comprehensive and effective support for agricultural production and decision-making. In the future, we will further improve the system's responses and the scope of questions it can address to meet the growing demand for agricultural information.

\subsubsection*{Acknowledgment.}This research project was supported in part by National Key Research and Development Program of China under Grant 2023YFF1000100, and in part by the Fundamental Research Funds for the Chinese Central Universities under Grant 2662023XXPY004.


\begin{thebibliography}{8}

\bibitem{ref_proc1}
GUO Wang, YANG Yusen, WU Huarui, ZHU Huaji, MIAO Yisheng, GU Jingqiu. Big Models in Agriculture: Key Technologies, Application and Future Directions[J]. Smart Agriculture, 2024, 6(2): 1-13.

\bibitem{ref_proc2}
Silva, B., Nunes, L., Estevão, R., \& Chandra, R. (2023). GPT-4 as an Agronomist Assistant? Answering Agriculture Exams Using Large Language Models. arXiv preprint arXiv:2310.06225.

\bibitem{ref_proc3}
Ye, S., Lauer, J., Zhou, M., Mathis, A., \& Mathis, M. (2024). AmadeusGPT: a natural language interface for interactive animal behavioral analysis. Advances in Neural Information Processing Systems, 36.

\bibitem{ref_proc4}
Achiam, J., Adler, S., Agarwal, S., Ahmad, L., Akkaya, I., Aleman, F. L., ... \& McGrew, B. (2023). Gpt-4 technical report. arXiv preprint arXiv:2303.08774.

\bibitem{ref_proc5}
Zhu, W., Han, R., Shang, X., Zhou, T., Liang, C., Qin, X.,  \& Li, L. (2024). The CropGPT project: Call for a global, coordinated effort in precision design breeding driven by AI using biological big data. Molecular Plant, 17(2), 215-218.

\bibitem{ref_proc6}
Devlin, J., Chang, M. W., Lee, K., \& Toutanova, K. (2018). Bert: Pre-training of deep bidirectional transformers for language understanding. arXiv preprint arXiv:1810.04805.
\bibitem{ref_proc7}
Yao, J. Y., Ning, K. P., Liu, Z. H., Ning, M. N., \& Yuan, L. (2023). Llm lies: Hallucinations are not bugs, but features as adversarial examples. arXiv preprint arXiv:2310.01469.
\bibitem{ref_proc8}
Patrick Lewis, Ethan Perez, Aleksandra Piktus, Fabio Petroni, Vladimir Karpukhin, Naman
Goyal, Heinrich Küttler, Mike Lewis, Wen-tau Yih, Tim Rocktäschel, et al. Retrieval-augmented
generation for knowledge-intensive nlp tasks. Advances in Neural Information Processing
Systems, 33:9459–9474, 2020.
\bibitem{ref_proc9}
Dong, J., Fatemi, B., Perozzi, B., Yang, L. F., \& Tsitsulin, A. (2024). Don't Forget to Connect! Improving RAG with Graph-based Reranking. arXiv preprint arXiv:2405.18414.
\bibitem{ref_proc10}
Yao, S., Zhao, J., Yu, D., Du, N., Shafran, I., Narasimhan, K., \& Cao, Y. (2022). React: Synergizing reasoning and acting in language models. arXiv preprint arXiv:2210.03629.
\bibitem{ref_proc11}
Wei, J., Wang, X., Schuurmans, D., Bosma, M., Xia, F., Chi, E., ... \& Zhou, D. (2022). Chain-of-thought prompting elicits reasoning in large language models. Advances in neural information processing systems, 35, 24824-24837./
\bibitem{ref_proc12}
Devlin, J., Chang, M. W., Lee, K., \& Toutanova, K. (2018). Bert: Pre-training of deep bidirectional transformers for language understanding. arXiv preprint arXiv:1810.04805.
\bibitem{ref_proc13}
Raffel, C., Shazeer, N., Roberts, A., Lee, K., Narang, S., Matena, M., ... \& Liu, P. J. (2020). Exploring the limits of transfer learning with a unified text-to-text transformer. Journal of machine learning research, 21(140), 1-67.

\bibitem{ref_proc14}
Lewis, M., Liu, Y., Goyal, N., Ghazvininejad, M., Mohamed, A., Levy, O., ... \& Zettlemoyer, L. (2019). Bart: Denoising sequence-to-sequence pre-training for natural language generation, translation, and comprehension. arXiv preprint arXiv:1910.13461.
\bibitem{ref_proc15}
Touvron, H., Lavril, T., Izacard, G., Martinet, X., Lachaux, M. A., Lacroix, T., ... \& Lample, G. (2023). Llama: Open and efficient foundation language models. arXiv preprint arXiv:2302.13971.

\bibitem{ref_proc16}
Brown, T., Mann, B., Ryder, N., Subbiah, M., Kaplan, J. D., Dhariwal, P., ... \& Amodei, D. (2020). Language models are few-shot learners. Advances in neural information processing systems, 33, 1877-1901.
\bibitem{ref_url1}
https://www.agri-chat.cn
\bibitem{ref_proc18}
Schick, T., Dwivedi-Yu, J., Dessì, R., Raileanu, R., Lomeli, M., Hambro, E., ... \& Scialom, T. (2024). Toolformer: Language models can teach themselves to use tools. Advances in Neural Information Processing Systems, 36.

\bibitem{ref_proc19}
Qing, J., Deng, X., Lan, Y., \& Li, Z. (2023). GPT-aided diagnosis on agricultural image based on a new light YOLOPC. Computers and Electronics in Agriculture, 213, 108168.

\bibitem{ref_proc20}
Balaguer, A., Benara, V., de Freitas Cunha, R. L., Estevão Filho, R. D. M., Hendry, T., Holstein, D., ... \& Chandra, R. (2024). RAG vs Fine-tuning: Pipelines, Tradeoffs, and a Case Study on Agriculture. arXiv e-prints, arXiv-2401.

\bibitem{ref_proc21}
Gao, Y., Xiong, Y., Gao, X., Jia, K., Pan, J., Bi, Y., ... \& Wang, H. (2023). Retrieval-augmented generation for large language models: A survey. arXiv preprint arXiv:2312.10997.

\bibitem{ref_proc22}
Shen, Y., Song, K., Tan, X., Li, D., Lu, W., \& Zhuang, Y. (2024). Hugginggpt: Solving ai tasks with chatgpt and its friends in hugging face. Advances in Neural Information Processing Systems, 36.
\bibitem{ref_proc23}
TANG Wentao, HU Zelin. Survey of Agricultural Knowledge Graph[J]. Computer Engineering and Applications, 2024, 60(2): 63-76.

\bibitem{ref_proc24}
Rezayi, S., Liu, Z., Wu, Z., Dhakal, C., Ge, B., Zhen, C., ... \& Li, S. (2022, July). AgriBERT: Knowledge-Infused Agricultural Language Models for Matching Food and Nutrition. In IJCAI (pp. 5150-5156).

\bibitem{ref_proc25}
Chen, Y., Kuang, J., Cheng, D., Zheng, J., Gao, M., \& Zhou, A. (2019). AgriKG: An agricultural knowledge graph and its applications. In Database Systems for Advanced Applications: DASFAA 2019 International Workshops: BDMS, BDQM, and GDMA, Chiang Mai, Thailand, April 22–25, 2019, Proceedings 24 (pp. 533-537). Springer International Publishing.



\end{thebibliography}
\end{document}